\documentclass[runningheads]{llncs}

\usepackage{eccv}

\usepackage{eccvabbrv}

\usepackage{graphicx}
\usepackage{booktabs}       

\usepackage[accsupp]{axessibility}  

\usepackage{hyperref}

\usepackage{orcidlink}

\usepackage[utf8]{inputenc} 
\usepackage[T1]{fontenc}    
\usepackage{hyperref}       
\usepackage{url}            
\usepackage{amsfonts}       
\usepackage{nicefrac}       
\usepackage{microtype}      
\usepackage{xcolor}         

\usepackage{enumitem}

\usepackage{amsmath}
\usepackage{multirow}
\usepackage{booktabs}
\usepackage{array}
\usepackage{caption}
\usepackage{subcaption}
\usepackage{adjustbox}
\usepackage[table]{xcolor}
\usepackage{marvosym}
\usepackage[normalem]{ulem}
\newcommand{\best}[1]{\textbf{#1}}
\newcommand{\second}[1]{\uline{#1}}

\usepackage{adjustbox}
\usepackage[table]{xcolor} 
\usepackage[normalem]{ulem} 

\newcommand{\chg}[1]{#1}

\begin{document}

\title{Progressive Pose-Guided 4D Animal Reconstruction from Monocular Video}


\author{Siyuan Li\inst{1} \orcidlink{0009-0001-3990-7502} \and
Weiying Chen\inst{1}\orcidlink{0009-0007-9212-0745} \and
Yilin Wang\inst{1}\orcidlink{0009-0004-3651-3614} \and \\
Xinxin Zuo\inst{2}\orcidlink{0000-0002-7116-9634} \and
Xingyu Li\inst{1}\orcidlink{0000-0002-3494-2552} \and
Li Cheng\inst{1}\orcidlink{0000-0003-3261-3533}\textsuperscript{\Letter}
}

\authorrunning{S.~Li et al.}

\institute{University of Alberta, Edmonton, AB, Canada \\
\email{\{sli20, lcheng5\}@ualberta.ca}\and
Concordia University, Montreal, QC, Canada\\
}

\maketitle

\begin{figure*}[htbp]
    \centering
    \includegraphics[width=1.\linewidth]{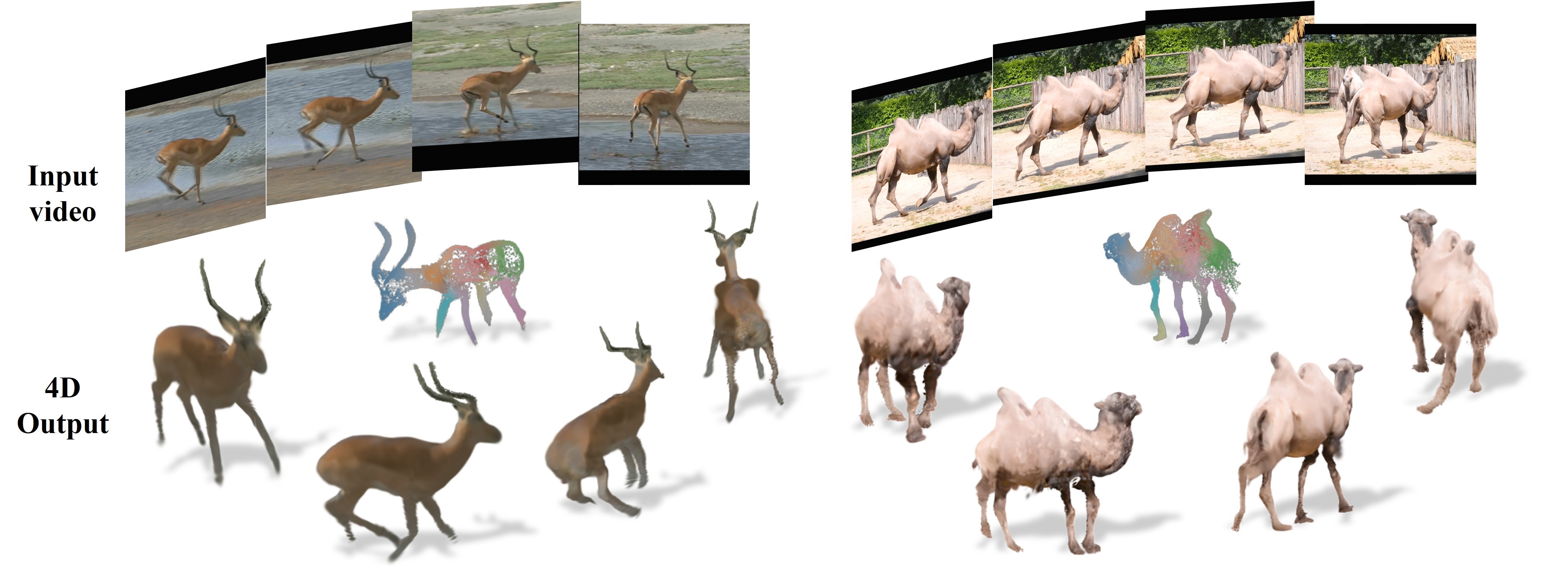}
    \caption{
    Given monocular videos of animals (top), our method produces 
    high-fidelity 4D models enabling free-viewpoint rendering 
    across time and viewing angles (bottom). Center: canonical 
    3D Gaussians colored by skinning weights.}

    \label{fig:teaser}
\end{figure*}

\begin{abstract}

Reconstructing 4D animals from monocular videos is challenging due to large inter-species variation, complex articulations, and the lack of reliable templates. Existing approaches 
typically rely on either strict category-specific priors that restrict generalization, or unconstrained generative models that sacrifice input fidelity. To bridge this gap, we present 
a progressive test-time optimization framework built on 3D Gaussian Splatting for high-fidelity 4D animal reconstruction from a single video. Our key insight is that a coarse shape prior suffices when coupled with a progressive strategy that disentangles articulated pose from non-rigid deformation. 
Specifically, we employ a symmetry-aware temporal encoding that exploits bilateral cues while absorbing camera estimation drift and a part-conditioned deformation mechanism guided by learnable part anchors and a learnable skinning field. 
Extensive experiments demonstrate that our approach generalizes robustly across diverse species, achieving superior geometric accuracy, temporal consistency, and visual fidelity compared to existing baselines, even under severe prior mismatch.
Project page: \url{https://syl-322.github.io/ReWild4D/}

\end{abstract}

\section{Introduction}
\label{sec:intro}

Animals in the natural world display a stunning diversity of shapes and behaviors. Accurately reconstructing their 3D shape and motion from visual data is crucial for various applications ranging from wildlife monitoring, animal conservation and ethology research, to immersive media content creation. 
Despite the wide accessibility of monocular video, the task of creating realistic 4D animal models from monocular video presents a significant challenge in computer vision. This is primarily due to the inherent complexity of animal morphology and behaviors, as well as the fact that their appearance and motion are only partly observable from a monocular video.

The task of 3D animal reconstruction presents unique challenges compared to 3D human reconstruction. 
Human models, such as SMPL \cite{loper2023smpl, SMPL-X:2019}, benefit from well-studied anatomical structures and abundant 3D motion capture datasets.  
In contrast, diverse animal species exhibit extreme shape and motion variations,
yet very little animal motion capture benchmarks are available. The pioneer SMAL \cite{zuffi2017smal} is a parametric SMPL-like animal model learned from a limited collection of toy figurines; it captures a rather limited category of species and lacks realistic details of shape and motion. The dilemma of both scarcely labeled, partially observable animal data, and extraordinarily diverse shape \& motion variations across animal species, forces a trade-off where, category-specific methods~\cite{badger2020birds3d, wang2021birds17, wu2023magicpony, rueegg2022barc, ruegg2023bite} achieve higher reconstruction quality by training on annotated dataset but struggle with generalization, while category-agnostic approaches~\cite{li2024fauna, aygun2024saor, jakab2024farm3d} improve coverage at the cost of reconstruction fidelity. 

Extending to 4D animal reconstruction from monocular videos reveals a fundamental tension between representation flexibility and computational efficiency. 
Mesh-based methods~\cite{yang2021lasr,yang2021viser,sabathier2024animalavatar} are efficient but topologically constrained, while neural implicit methods~\cite{yang2022banmo,yang2023rac} offer flexibility at prohibitive computational costs. 
3D Gaussian Splatting (3DGS)~\cite{kerbl20233dgs} emerges as a powerful alternative: it combines the rendering efficiency of explicit geometry with the topological flexibility of unstructured point-based models, making it uniquely suited to capture complex dynamics.
However, recent 3DGS-based frameworks for animal reconstruction, such as GART~\cite{lei2024gart}, still rely on parametric templates.
Generation-based methods~\cite{ren2024l4gm,zhang2025gvfdiffusion} achieve impressive synthesis but sacrifice input fidelity. 
Existing methods either depend on rigid templates that limit generalization or generative priors that compromise reconstruction accuracy. We argue this trade-off stems from treating shape priors as strict constraints rather than flexible initializations. 

Inspired by the above observations, we present a progressive test-time optimization framework built on 3D Gaussian Splatting for high-fidelity 4D animal reconstruction from monocular in-the-wild videos, with representative results shown in \cref{fig:teaser}.
Our key insight is that robust 4D reconstruction does not require highly accurate shape priors; a coarse initialization suffices when allowed to continuously evolve through a principled disentanglement of articulated pose and non-rigid deformation.
To achieve this, we introduce a bilateral symmetry augmentation strategy that exploits mirror cues while absorbing systematic camera drift. Furthermore, rather than isolating pose and deformation stages like prior pipelines, our framework seamlessly bridges them using learnable part anchors. Combined with a symmetry-aware temporal encoding and part-conditioned cross-attention, these anchors provide a shared identity space that intrinsically guides local geometry updates.
This progressive disentanglement enables temporally coherent reconstructions across diverse species and complex behaviors.
In summary, our approach features the following key contributions:
\begin{itemize}
\item A progressive test-time optimization framework for 4D animal reconstruction from monocular video, requiring neither annotated training data, category-specific templates, nor multi-view generative priors. By treating the shape prior as a coarse initialization with a learnable skinning field, our method generalizes across diverse species even under severe prior mismatch.

\item A two-stage pipeline that disentangles articulated 
motion from non-rigid deformation. For pose refinement, we introduce a symmetry-aware temporal encoding that explicitly decouples 2D visual correspondence from 3D camera estimation drift. For non-rigid deformation, we propose a part-conditioned cross-attention mechanism, where learnable part anchors serve a dual role: generating part-level temporal features for articulated pose estimation, and providing per-Gaussian identity embeddings to spatially condition local deformations.

\item Extensive experiments demonstrate state-of-the-art 
performance in geometric accuracy, temporal consistency, 
and visual fidelity across diverse species on challenging 
in-the-wild videos. 
\end{itemize}

\section{Related Work}
\label{sec:literature}

\subsubsection{3D Animal Reconstruction.}
Reconstructing 3D animals is more challenging than reconstructing humans due to interspecific variation, complex articulations, and limited 3D data. 
Parametric models such as SMAL\cite{zuffi2017smal} provide the first skinned multi-animal template from toy figurines, followed by various extensions and refinement\cite{zuffi2018smalr,Zuffi2019smalst, rueegg2022barc,ruegg2023bite}.
CSM-based methods\cite{kulkarni2019csm,kulkarni2020acsm} predict dense image-to-surface mappings but remain tied to predefined templates. 
Template-free methods\cite{yao2022lassie, yao2023hilassie, liu2023lepard} discover parts and skeletons from sparse images through optimizing primitive part representation.
Recent learning-based approaches scale to Internet data: UMR\cite{li2020umr}, MagicPony\cite{wu2023magicpony}, and Farm3D\cite{jakab2024farm3d} learn category-specific models, whereas FAUNA\cite{li2024fauna} and SAOR\cite{aygun2024saor} aim for category-agnostic reconstruction.
The evolution in 3D animal priors has grounded a natural basis for 4D reconstruction, yet they are often treated as fixed constraints, and are severely limited in generalizing to unseen animal species.
Instead, a shape prior is engaged in our approach as merely coarse initialization, which has been empirically demonstrated to contribute to flexible and faithful 4D recovery across species.

\subsubsection{Dynamic Animal Reconstruction}
Extending static 3D models to capture temporal dynamics from monocular videos remains a central challenge in animal reconstruction. 
Deformation-based approaches\cite{yang2021lasr,yang2021viser,wu2022casa} rely on deforming initial meshes with fixed connectivity, which struggle to recover fine surface details. 
~\cite{yang2022banmo,yang2023rac} adopt NeRF-based representations for greater topological flexibility but suffer from prohibitive computational costs and lack explicit surface geometry.
Hybrid explicit approaches\cite{sabathier2024animalavatar, lei2024gart} have recently emerged, among which GART\cite{lei2024gart} leverages 3D Gaussian Splatting for a better efficiency-flexibility trade-off; however, these pipelines typically treat pose and deformation as isolated stages. This decoupling limits their ability to re-bind parts coherently under severe prior mismatches. In contrast, our dual-role part anchors establish a shared identity space that seamlessly couples both stages.

\subsubsection{Gaussian Splatting for Dynamic Scenes}
3D Gaussian Splatting (3DGS)~\cite{kerbl20233dgs} revolutionizes novel view synthesis by 
providing an explicit, flexible representation that achieves real-time performance with high fidelity. Extending it to dynamic scenes, current variants generally fall into two categories.
\emph{Time-augmented} 3DGS encodes temporality in Gaussians but scales poorly with sequence length\cite{wu20244dgs,xu2024grid4d,oh2025hybrid4dgs}. 
\emph{Deformation-based} variants keep a canonical 3DGS and learn warps for compact memory and smooth motion\cite{yang2024deform3dgs,huang2024sc,song2025spline}. We adopt the deformation paradigm and drive warps with pose cues to improve temporal coherence for our task.

\subsubsection{Video-to-4D Generation}
A parallel line of work leverages generative priors for video-to-4D synthesis. 
Zero-1-to-3\cite{liu2023zero123} pioneered this direction by leveraging diffusion models to hallucinate novel views from a single image. \chg{ Building upon Zero-1-to-3, DreamMesh4D\cite{li2024dreammesh4d} generates 4D models using a hybrid mesh-3DGS representation. In contrast, PAD3R\cite{pad3r} reconstructs 4D content based on Zero-1-to-3 by training a personalized PoseNet.}
SV4D\cite{xiesv4d} and SV4D 2.0\cite{yao2025sv4d2} extend this concept to videos, enforcing multi-frame and multi-view consistency. 
However, these methods struggle with long video sequences, often losing geometric detail and requiring fixed-length inputs.
Other methods that directly generate 4D models, such as Splat4D\cite{yin2025splat4d}, L4GM\cite{ren2024l4gm}, and GVF-Diffusion\cite{zhang2025gvfdiffusion} usually demand large model ensembles and high memory usage. They will also produce results with significant inconsistencies in both appearance and shape compared to the input video.
A fundamental limitation of these generative approaches is that they typically freeze the canonical representation once dynamic learning begins; consequently, their canonical space becomes biased toward the dataset distribution rather than the specific input subject. In contrast, we pursue reconstruction-only supervision from the input video and allow the canonical shape and skinning field to continue evolving, avoiding generative mismatch while retaining strict faithfulness to the input identity.

\begin{figure}[tb]
  \centering
    \includegraphics[width=1.0\linewidth]{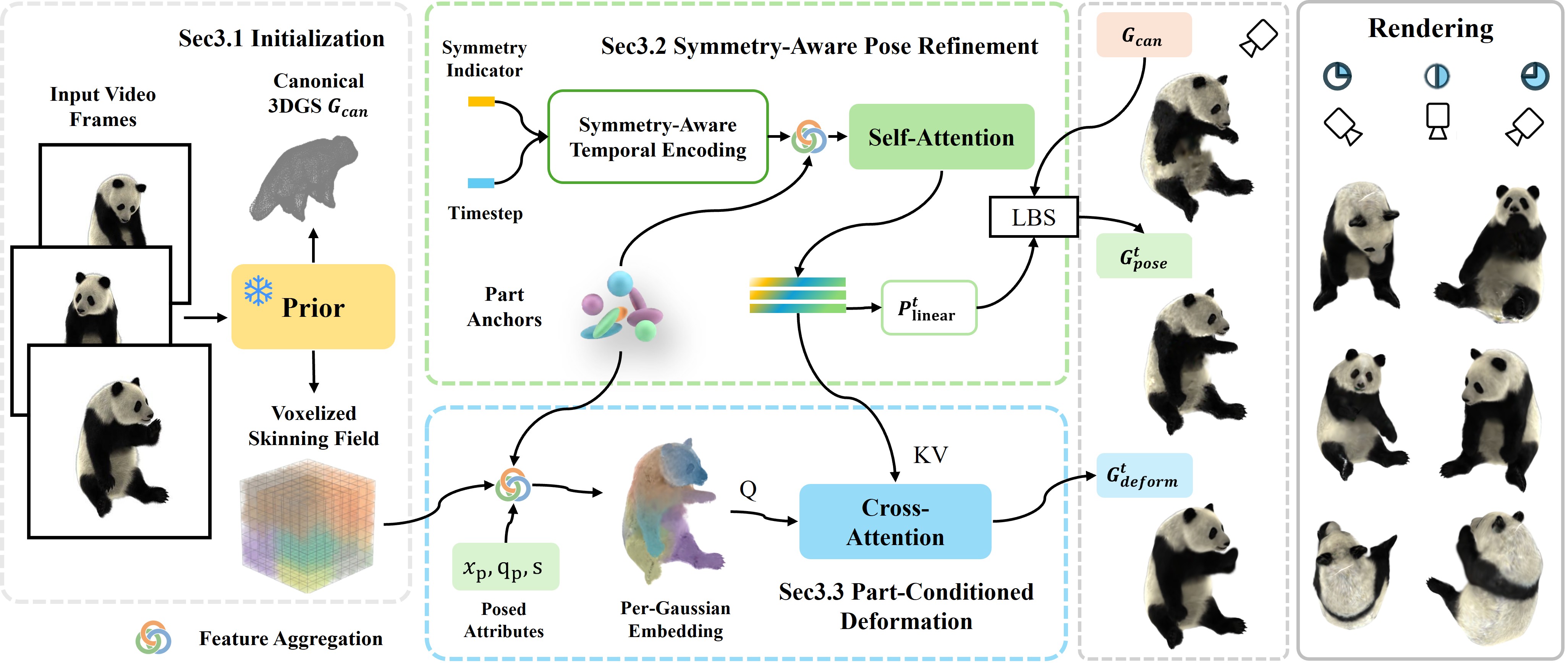}
    \caption{
    \textbf{Pipeline overview.} From a monocular video, we initialize 
    canonical 3D Gaussians and a learnable skinning field from the 
    Fauna prior (Sec.~\ref{sec:prior}). Symmetry-Aware Pose Refinement 
    (Sec.~\ref{sec:pose}): learnable part anchors and symmetry-aware 
    temporal encoding are processed by self-attention to estimate 
    per-joint transformations, yielding intermediate representations $G_{\text{pose}}^t$. 
    Part-Conditioned Deformation (Sec.~\ref{sec:deform}): 
    part anchors are aggregated via sampled skinning weights into per-Gaussian 
    part embeddings and, together with posed attributes, query the 
    part-level temporal features through cross-attention to produce 
    $G_{\text{deform}}^t$. Notice how the deformation module captures fine-grained dynamics (e.g., the raised front paw in $G^t_{\text{deform}}$) that are completely missed by the rigid LBS in $G^t_{\text{pose}}$. Right: rendering from arbitrary viewpoints and time steps.
    }
    \label{fig:pipeline}
\end{figure}

\section{Our Approach}
\label{sec:method}

Our goal is to recover high-fidelity, time-varying 4D representations of animals from a monocular video sequence $\{I^t\}_{t=1}^T$. 
To address the inherent ambiguity of monocular observations, we employ a progressive test-time optimization framework that explicitly disentangles articulated pose from non-rigid deformation. 
Specifically, initialized from a coarse prior and organized under a bilateral symmetry augmentation strategy (\cref{sec:prior}), 
the canonical 3D Gaussian model $G_{\text{can}}$ first undergoes pose refinement driven by learnable part anchors and symmetry-aware temporal encoding (\cref{sec:pose}), followed by part-conditioned deformation (\cref{sec:deform}).
The complete pipeline is illustrated in \cref{fig:pipeline}.

\subsection{Initialization and Bilateral Symmetry Augmentation}
\label{sec:prior}

\subsubsection{Initialization}
We bootstrap using Fauna~\cite{li2024fauna}, which provides a coarse estimate of animal shape and pose from a single image. 
Given an input frame $I^t$, Fauna predicts 
rest-pose vertices $V^t \in \mathbb{R}^{N_v \times 3}$, 
skinning weights $W^t \in \mathbb{R}^{N_v \times J}$, 
camera poses $C^t \in \mathbb{R}^{4 \times 4}$, and 
joint transformations $P^t \in \mathbb{R}^{J \times 4 \times 4}$. 
We retain the first-frame outputs $(V^1, W^1)$ for canonical initialization, and use $\{C^t\}$ and $\{P^t\}$ as starting points for optimization.

Following~\cite{kerbl20233dgs}, we represent the canonical animal as 3D Gaussians 
$G_{\text{can}}=\{\mathbf{x}, \mathbf{q}, \mathbf{s}, \boldsymbol{\alpha}, \mathbf{c}\}$, initialized from $V^1$. 
We embed $W^1$ into a dense voxel grid 
$W_g \in \mathbb{R}^{J \times 32 \times 32 \times 32}$ 
inspired by~\cite{jiang2022selfrecon, lei2024gart}. Since shape priors are often geometrically misaligned with the target video, we formulate $W_g$ as a \textit{learnable} skinning field optimized alongside the Gaussians, enabling continuous interpolation for dynamically spawned points and progressive correction of initial binding inaccuracies. Each Gaussian's skinning weights $\mathbf{w}_i \in \mathbb{R}^J$ are obtained via trilinear interpolation from $W_g$.

\subsubsection{Bilateral Symmetry Augmentation}
We assume the canonical shape is bilaterally symmetric and that $\{I^t, I^t_{\text{flip}}\}$ are mirror pairs; thus, horizontally flipping can extend viewpoint coverage. Running the prior on both yields 
$\{C^t, C^t_{\text{flip}}\}$ that \emph{should} satisfy $C^t_{\text{flip}}{=}MC^t$ ($M$: reflection transformation along bilateral symmetry plane). 
In practice, learning-based estimators often exhibit a systematic flip-induced bias: $C^t_{\text{flip}}{\neq}MC^t$. 

Naively fitting both as exact mirrors injects this drift as conflicting supervision, while treating them as independent samples discards their inherent geometric relationship. 
To leverage symmetry without introducing noise or missing geometric context, 
we directly apply $M$ to $C^t$ and $C^t_{\text{flip}}$, 
forming two internally-symmetric camera groups $\{C^t, C^t_{\text{sym}}\}$ and $\{C^t_{\text{flip}}, C^t_{\text{flip,sym}}\}$.
Thus, we can construct two complementary supervision groups:
\begin{align}
\mathcal{V}_{\text{orig}} &= \{(I^t, C^t, P^t),\; 
(I^t_{\text{flip}}, C^t_{\text{sym}}, P^t_{\text{flip}})\}, \\
\mathcal{V}_{\text{flip}} &= \{(I^t_{\text{flip}}, C^t_{\text{flip}}, 
P^t_{\text{flip}}),\; (I^t, C^t_{\text{flip,sym}}, P^t)\},
\end{align}
each containing an original-flipped frame pair under a shared camera
reference. 
Each training iteration samples from either 
$\mathcal{V}_{\text{orig}}$ or $\mathcal{V}_{\text{flip}}$, 
enriching supervision by exposing the model to both geometric 
interpretations of the same scene.
A symmetry-aware temporal encoding then lets the network distinguish input samples during pose refinement (Sec.~\ref{sec:pose}).

\subsection{Symmetry-Aware Pose Refinement}
\label{sec:pose}
Initial per-frame poses $P^t$ predicted by the prior model are often unreliable due to limited views and articulation ambiguity. Our pose refinement module estimates per-joint transformations for linear blend skinning (LBS), mapping the canonical 
representation $G_{\text{can}}$ to posed states $G^t_{\text{pose}}$.

To incorporate bilateral symmetry and handle the systematic camera drift introduced above, we design a \textbf{symmetry-aware temporal encoding}:
\begin{align}
\mathbf{e}_{m,v}^{t} = \text{emb}(t \cdot m) \oplus v, \quad m,v \in \{-1,1\},
\label{eq:encode}
\end{align}
where $\oplus$ denotes concatenation. 
The indicator $v$ encodes the 2D frame state (1 for $I^t$ and -1 for $I^t_\text{flip}$), while $m$ identifies the 3D camera reference (1 for $\mathcal{V}_{\text{orig}}$ and -1 for $\mathcal{V}_{\text{flip}}$). 
This factorization decouples 2D visual symmetry from 3D camera configuration, allowing the model to exploit mirror relationships while absorbing inter-group camera drift rather than 
treating it as conflicting supervision.
Multiplying $t$ by $m$ before encoding further leverages the antisymmetric property of sinusoidal embeddings to structurally relate the two camera groups.

Instead of predicting joint poses solely from a temporal embedding, we introduce randomly initialized learnable part anchors $\mathbf{A} \in \mathbb{R}^{J \times D}$ that learn joint-specific identity representations shared across all frames. We set $J=20$ following the Fauna prior and $D=8$ for the anchor dimension.
The anchors $\mathbf{A}$ combined with $\mathbf{e}_{m,v}^t$ are processed by a self-attention block to produce joint-specific temporal features $\mathbf{F}_J^t \in \mathbb{R}^{J \times K}$. 
These features are projected to per-joint transformations $P^t_{\text{linear}} \in \mathbb{R}^{J \times 7}$ (quaternion rotation and translation), and also forwarded to the subsequent 
deformation stage as kinematic context. 
Applying LBS with $P^t_{\text{linear}}$ and interpolated skinning weights updates Gaussian centers and orientations, yielding the posed Gaussians $G^t_{\text{pose}}= \left\{\mathbf{x}_{\text{p}}, \mathbf{q}_{\text{p}}, \mathbf{s}, \boldsymbol{\alpha}, \mathbf{c} \right\}$.

To prevent appearance gradients from corrupting pose estimation, we supervise this stage with silhouette loss only, detaching non-articulation attributes during rendering (\cref{sec:optim}). This gradient isolation ensures that pose refinement is driven purely by geometric alignment.

\subsection{Part-Conditioned Deformation}
\label{sec:deform}
While LBS-based pose refinement captures articulated motion, real animals exhibit non-rigid effects that cannot be modeled by skeletal transformations alone. Our deformation module 
addresses this by predicting the final deformed geometric parameters $G^t_{\text{deform}} = \{\mathbf{x}_{\text{d}}, \mathbf{q}_{\text{d}}, \mathbf{s}_{\text{d}}, \boldsymbol{\alpha}, \mathbf{c}\}$, conditioned on the articulation context from pose refinement.

To provide explicit kinematic context beyond spatial position, we augment the query with a part-aware embedding $\mathbf{z}_i = \mathbf{w}_i^\top \mathbf{A} \in \mathbb{R}^D$, which encodes each Gaussian's joint membership through the \emph{same} learnable anchors $\mathbf{A}$ used in pose refinement (\cref{sec:pose}), making $\mathbf{A}$ a shared identity space that couples the two stages.
This complements the positional encoding by providing a position-independent part identity signal, improving robustness when different body parts overlap in posed space.

The query for cross-attention is constructed from $\mathbf{z}_i$ and the posed geometric attributes of each Gaussian $(\mathbf{x}_{\text{p},i}, \mathbf{q}_{\text{p},i}, \mathbf{s}_i)$, while the joint-specific temporal features $\mathbf{F}_J^t$ from pose refinement serve as keys and values. 
This enables each Gaussian to selectively retrieve time-varying deformation signals from its associated joints, guided by both spatial proximity and learned part membership.
The network outputs only the deformed geometry $(\mathbf{x}_{\text{d}}, \mathbf{q}_{\text{d}}, \mathbf{s}_{\text{d}})$; the appearance $(\boldsymbol{\alpha}, \mathbf{c})$ is not predicted per-frame but shared from the canonical Gaussians and optimized directly, maintaining appearance consistency across time.

\begin{figure}[!t]
  \centering
    \includegraphics[width=1.0\linewidth]{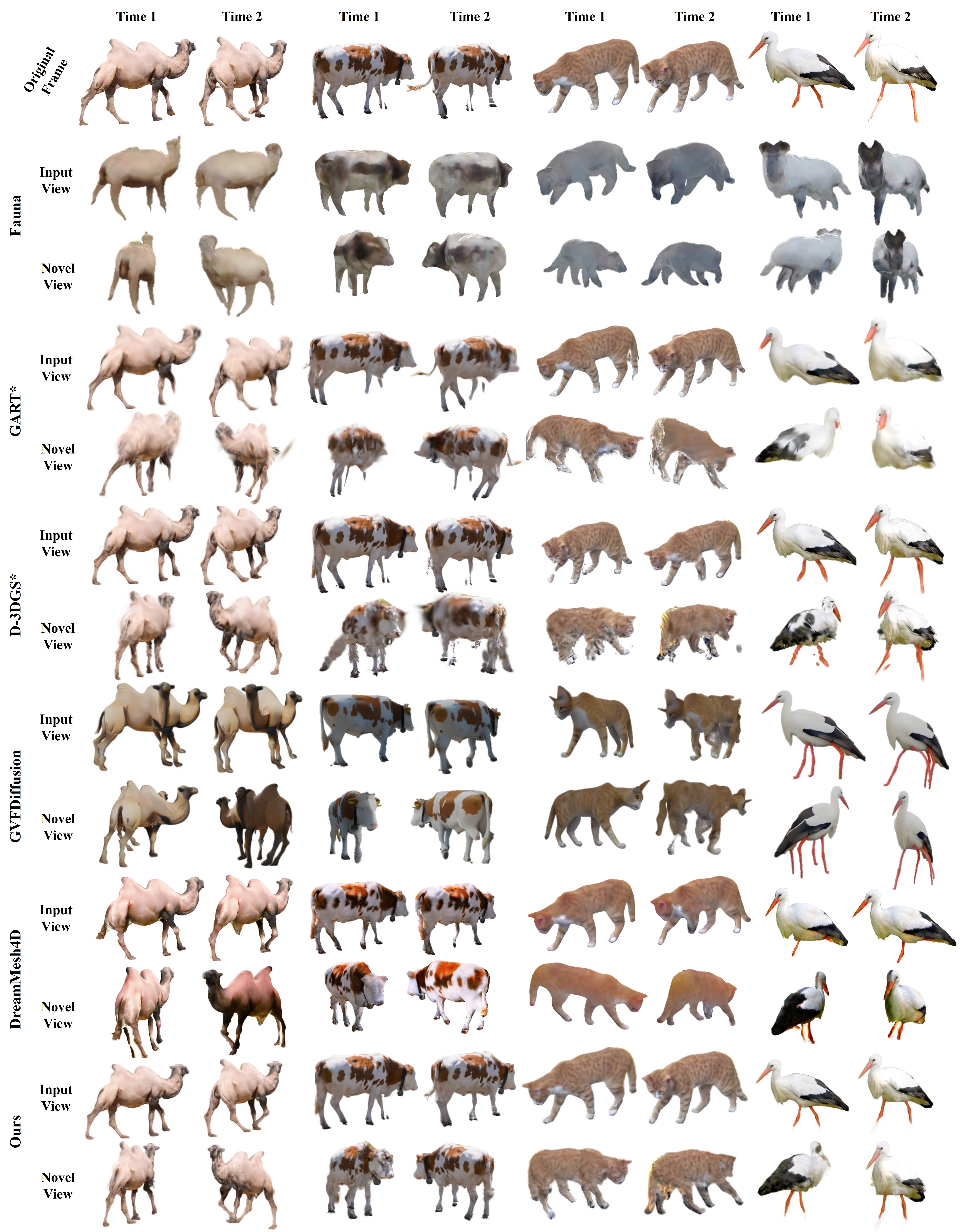}
    \caption{Visual comparison with SOTA methods on real videos at 2 different time steps.}
    \label{fig:compare}
\end{figure}

\subsection{Progressive Optimization and Objectives}
\label{sec:optim}

We optimize our framework through progressive test-time optimization on each monocular video. The overall objective balances geometric articulation and non-rigid shape fidelity:
\begin{equation}
\mathcal{L}^t_{\text{total}} = \mathcal{L}^t_{\text{pose}} + \mathcal{L}^t_{\text{deform}} + \mathcal{L}^t_{\text{geo}},
\end{equation}
where $\mathcal{L}^t_{\text{pose}}$ isolates articulated motion, $\mathcal{L}^t_{\text{deform}}$ captures fine details, and $\mathcal{L}^t_{\text{geo}}$ provides geometric regularization.

\subsubsection{Stage-Aware Rendering Losses.}
To prevent appearance gradients from corrupting geometric 
articulation, the pose refinement stage relies strictly on 
silhouette supervision. We block appearance-driven gradients 
by detaching non-articulated attributes during rasterization, 
yielding the posed Gaussians 
$G^t_{\text{pose}}=\{\mathbf{x}_{\text{p}}, 
\mathbf{q}_{\text{p}}, 
\text{sg}(\mathbf{s}, \boldsymbol{\alpha}, \mathbf{c})\}$. 
The pose loss is:
\begin{equation}
\mathcal{L}^t_{\text{pose}} = \lambda_{\text{pose}} \, 
\mathcal{L}_{\text{sil}}(\hat{S}^t_{\text{pose}}, 
S^t_{\text{SAM}}),
\end{equation}
where $\mathcal{L}_{\text{sil}}$ combines standard BCE and 
Dice losses, $\hat{S}^t_{\text{pose}}$ is the rendered 
silhouette, $S^t_{\text{SAM}}$ is the Grounded-SAM 
\cite{ren2024grounded} mask, and $\lambda_{\text{pose}}=0.2$.

Conversely, the deformation stage jointly optimizes all Gaussian attributes (including appearance) to capture non-rigid dynamics and realistic textures. The deformation loss combines silhouette and photometric objectives:
\begin{equation}
\mathcal{L}^t_{\text{deform}} = \mathcal{L}_{\text{rgb}}(
\hat{I}^t_{\text{deform}}, I^t) + \lambda_{\text{sil}} \, 
\mathcal{L}_{\text{sil}}(\hat{S}^t_{\text{deform}}, 
S^t_{\text{SAM}}),
\end{equation}
where $\mathcal{L}_{\text{rgb}}$ is a standard blend of $\mathcal{L}_1$ and SSIM losses ($\lambda_{\text{SSIM}}=0.2$), $\hat{I}^t_{\text{deform}}$ is the rendered RGB, $I^t$ is the input frame, and $\lambda_{\text{sil}}=0.1$.

\subsubsection{Geometric Regularization.}

Our geometric regularization term $\mathcal{L}_{\text{geo}}$ consists of three components: a skinning field smoothness term $\mathcal{L}_{\text{tv}}$, a part-aware compactness term $\mathcal{L}_{\text{compact}}$, and a surface normal smoothness term $\mathcal{L}_{\text{smooth}}$:
\begin{equation}
\mathcal{L}^t_{\text{geo}} = \lambda_{\text{tv}}\,
\mathcal{L}_{\text{tv}} + \lambda_{\text{compact}}\,
\mathcal{L}_{\text{compact}} + \lambda_{\text{smooth}}\,
\mathcal{L}^t_{\text{smooth}},
\end{equation}
where $\lambda_{\text{tv}}=10$, $\lambda_{\text{compact}}=0.1$, and $\lambda_{\text{smooth}}=0.1$. $\mathcal{L}^t_{\text{geo}}$is only applied during the first 10k iterations. See details in the supplementary material.

First, to ensure physically plausible skinning fields, we apply a Huber total variation loss $\mathcal{L}_{\text{tv}}$ on the skinning weights in the voxel grid $W_g \in \mathbb{R}^{J \times R \times R \times R}$:
\begin{equation}
\mathcal{L}_{\text{tv}} = \frac{1}{3} \sum_{d \in \{x,y,z\}} \mathbb{E} \left[ \mathcal{H}_{\delta=0.05} (\nabla_d W_{g}) \right],
\end{equation}
where $\nabla_d$ denotes spatial gradients along axis $d$ and $\mathcal{H}_\delta$ is the Huber function.

Second, to prevent floating artifacts, we introduce a part-aware compactness loss applied to both canonical and posed Gaussian positions. For each joint $j$, we compute the weighted centroid and covariance using the current skinning weights $\{w_{i,j}\}$, and penalize only the two smallest eigenvalues $\lambda_{j,1} \leq \lambda_{j,2}$ of the covariance, preserving elongated structures (e.g., limbs) along the principal axis while discouraging lateral spread:
\begin{equation}
\mathcal{L}_{\text{compact}} = \frac{1}{J} \sum_{j=1}^{J} 
\gamma_j (\lambda_{j,1} + \lambda_{j,2}),
\end{equation}
where $\gamma_j = \exp(-\alpha \cdot \Omega_j / \max_k \Omega_k)$ down-weights large parts such as the torso, $\alpha$ is a scaling hyperparameter (we set $\alpha=2.0$) and $\Omega_j {=} \sum_i w_{i,j}$ is the occupancy mass of joint $j$.

Third, to reduce surface noise in novel views, we render normal maps $\hat{N}^t_\theta$ from randomly sampled viewpoints $\theta$ and penalize local angular variation using absolute cosine similarity to handle normal orientation ambiguity:
\begin{equation}
\mathcal{L}^t_{\text{smooth}} = \sum_{d\in\{x,y\}} 
\frac{\sum_{p} \omega^t_{p,d} \bigl(1 - |\hat{N}^t_\theta(p) 
\cdot \hat{N}^t_\theta(p+\delta_d)|\bigr)}
{\sum_{p} \omega^t_{p,d} + \epsilon},
\end{equation}
where $\delta_d$ denotes unit pixel offsets along each axis and 
$\omega^t_{p,d} = \hat{S}^t_\theta(p) \cdot \hat{S}^t_\theta(p+\delta_d)$ 
restricts computation to foreground regions.

\subsubsection{Progressive Stabilization.}
To prevent overfitting to inaccurate priors, the predicted pose $\hat{\mathbf{P}}^t_{\text{linear}}$ is blended with the Fauna prior using a weight that anneals from 1 to 0 over 7K iterations, and an $L_2$ regularization toward the prior is applied during the first 4K iterations. Both constraints vanish during training, ensuring the final reconstruction is not limited by prior accuracy. Following~\cite{yang2024deform3dgs}, 
we also inject decaying Gaussian noise into the temporal coordinate $t$ during early iterations to encourage temporal smoothness.

\begin{table*}[tb]
\centering
    \begin{minipage}{1.0\linewidth}
    \centering
    \caption{\textbf{Input-view} quality on three datasets. Best in \textbf{bold}, second best \uline{underlined}. }
    \label{tab:input-views}
    \setlength{\tabcolsep}{6pt}
    \renewcommand{\arraystretch}{1.15}
    \adjustbox{max width=\columnwidth}{
        \begin{tabular}{lccccccccc}
        \toprule
        & \multicolumn{3}{c}{\textbf{DAVIS}} & \multicolumn{3}{c}{\textbf{Online}} & \multicolumn{3}{c}{\textbf{APTv2}} \\
        \cmidrule(lr){2-4}\cmidrule(lr){5-7}\cmidrule(lr){8-10}
        \textbf{Method} & PSNR$\uparrow$ & SSIM$\uparrow$ & LPIPS$\downarrow$
                       & PSNR$\uparrow$ & SSIM$\uparrow$ & LPIPS$\downarrow$
                       & PSNR$\uparrow$ & SSIM$\uparrow$ & LPIPS$\downarrow$ \\
        \midrule
        Fauna \cite{li2024fauna}         & 18.281 & 0.761 & 0.280 & 16.669 & 0.774 & 0.279 & 19.561 & 0.760 & 0.267 \\
        D-3DGS* \cite{yang2024deform3dgs}        & \chg{\second{25.776}}& \chg{\second{0.905}} & \chg{\second{0.100}} & \chg{\second{25.659}} & \chg{\second{0.912}} & \chg{\second{0.096}} & \chg{23.636} & \chg{0.850} & \chg{0.144} \\
        GART \cite{lei2024gart} & 19.486 & 0.810 & 0.201 & 21.006 & 0.841 & 0.181 & 18.564 & 0.807 & 0.168 \\
        GART* \cite{lei2024gart}  & 21.347 & 0.859 & 0.171 & {23.128} & 0.883 & 0.158 & 19.167 & 0.834 & {0.150} \\
        GVFDiffusion \cite{zhang2025gvfdiffusion}   & 16.419 & 0.836 & 0.174 & 16.835 & 0.857 & 0.167 & 14.820 & 0.778 & 0.256 \\
        \chg{DreamMesh4D} \cite{li2024dreammesh4d}   & \chg{23.150} & \chg{0.881} & \chg{0.116} & \chg{23.859} & \chg{0.889} & \chg{0.131} & \chg{\best{26.000}} & \chg{\best{0.899}} & \chg{\best{0.102}} \\
        \rowcolor{gray!15} 
        Ours    & \best{26.321} & \best{0.920} & \best{0.089} & \best{26.206} & \best{0.923} & \best{0.091} \ & \second{24.938} & \second{0.878} & \second{0.139} \\
        \bottomrule
        \end{tabular}}
    \end{minipage}

    \vspace{1em}
    \begin{minipage}{1.0\linewidth}
    \centering
    \caption{\textbf{Novel-view} quality on three datasets. Best in \textbf{bold}, second best \uline{underlined}. A dash indicates metric not reported by the method.}
    \label{tab:novel-views}
    \setlength{\tabcolsep}{6pt}
    \renewcommand{\arraystretch}{1.15}
    \adjustbox{max width=\columnwidth}{
        \begin{tabular}{lccc ccc ccc}
        \toprule
        & \multicolumn{3}{c}{\textbf{DAVIS}} & \multicolumn{3}{c}{\textbf{Online}} & \multicolumn{3}{c}{\textbf{APTv2}} \\
        \cmidrule(lr){2-4}\cmidrule(lr){5-7}\cmidrule(lr){8-10}
        \textbf{Method} & KID-16V$\downarrow$ & FVD-F$\downarrow$ & FVD-Diag$\downarrow$
                        & KID-16V$\downarrow$ & FVD-F$\downarrow$ & FVD-Diag$\downarrow$
                        & KID-16V$\downarrow$ & FVD-F$\downarrow$ & FVD-Diag$\downarrow$ \\
        \midrule
        Fauna \cite{li2024fauna}           & 0.279 & {---} & --- & 0.334 & --- & --- & 0.247 & --- & --- \\
        D-3DGS* \cite{yang2024deform3dgs}        & \chg{0.199} & \chg{1192.015} & \chg{\second{980.747}} & \chg{0.231} & \chg{\second{1244.444}} & \chg{\second{1268.110}} & \chg{0.262} & \chg{1181.275} & \chg{1112.902} \\
        GART \cite{lei2024gart} & 0.216 & 1750.899 & 1675.388 & 0.233 & 1470.950 & 1547.561 & 0.230 & 1355.4969 & 992.195 \\
        GART* \cite{lei2024gart}  & 0.208 & 1680.705 & 1579.871 & 0.238 & 1473.927 & 1364.587 & {0.228} & {1134.661} & {948.274} \\
        GVFDiffusion \cite{zhang2025gvfdiffusion}   & \second{0.145} & 1872.192 & 1270.189 & \second{0.179} & 1673.419 & 1387.732 & 0.274 & 1715.471 & 1575.461 \\
        \chg{DreamMesh4D \cite{li2024dreammesh4d}}    & \chg{0.148} & \chg{\second{1154.038}} & \chg{2257.034}  & \chg{{0.185}} & \chg{{1518.163}} & \chg{{2911.299}} & \chg{\best{0.188}} & \chg{\best{629.482}} & \chg{\best{720.280}}\\
        \rowcolor{gray!15} 
        
        Ours    & \best{0.108} & \best{895.708} & \best{696.166} & \best{0.160} &  \best{972.028} & \best{1245.358} & \second{0.189} & \second{996.336} & \second{916.325} \\
        \bottomrule
        \end{tabular}}
    \end{minipage}
\end{table*}

\section{Experiment}
\label{sec:exp}

\subsection{Dataset}

We collect 87 in-the-wild videos from three sources: online collection (11 videos), DAVIS~\cite{Perazzi2016davis} (8 videos), and APTv2~\cite{yang2023aptv2} (68 videos). For APTv2, all videos contain 15 frames, except for two manually composed by concatenating similar clips. We extract animal masks using source-specific strategies (details in supplementary material), estimate animal and camera parameters for both original and horizontally flipped sequences using Fauna, and select temporally stable frames via DBSCAN clustering~\cite{ester1996dbscan} on camera trajectories, resulting in a 4:1 train/test split. 

To obtain more reliable measurements, we further evaluate on the Artemis dataset~\cite{luo2022artemis}, which provides 9 CGI animal species with 24 motions captured by 36/72 synchronized cameras. We optimize on a single sampled monocular sequence and test on the remaining held-out views (1:71 for wolf, 1:35 for others) with dataset-provided cameras.

\subsection{Implementation Details}
We implement our method in PyTorch. Optimization runs for 20K iterations on DAVIS, Online and Artemis, and 10K on APTv2. For in-the-wild videos, we apply pose blending and prior regularization as described in \cref{sec:optim}; for Artemis, these are disabled. Further details are in the supplementary material. On a single NVIDIA A6000 GPU at $512 \times 512$ resolution, rendering averages 105 FPS, and per-video optimization takes approximately 13 minutes on APTv2 (10K iterations).

\subsection{Results and Comparisons}

\subsubsection{Baselines}

We compare against five methods across two paradigms. 
Reconstruction-based: (1) Fauna\cite{li2024fauna}, a single-image 3D model serving as our initialization prior; (2) GART\cite{lei2024gart}, an articulated monocular optimization framework evaluated both standardly and with test-time refinement (denoted as GART*); and (3) D-3DGS\cite{yang2024deform3dgs}, a dynamic 3DGS approach. For strict fairness, we equip D-3DGS with our Fauna initialization and silhouette supervision (denoted D-3DGS*). Additionally, we enhance both GART and D-3DGS with flip augmentation (assigning timestamp $t \cdot v$ to flipped frames). 
Generation-based: Video-to-4D models (4) GVFDiffusion\cite{zhang2025gvfdiffusion} and (5) DreamMesh4D\cite{li2024dreammesh4d}, powered by heavy pre-trained priors Trellis\cite{xiang2025trellis} and Zero-1-to-3\cite{liu2023zero123}, respectively.

\begin{table}[tb]
\centering
\caption{
Quantitative results on the Artemis dataset, covering method comparison and component ablations.
Best in bold, second best \underline{underlined}.
}
\vspace{-3mm}
    \begin{subtable}[t]{0.54\linewidth}
        \centering
        \caption{Novel-view quality and 3D geometric accuracy on the Artemis dataset.}
        \label{tab:artemis_compare}
        \vspace{-7pt} %
        \renewcommand{\arraystretch}{1.32} 
        \resizebox{\linewidth}{!}{
            \begin{tabular}{lccccccc}
            \toprule
            \multirow{2}{*}{\textbf{Methods}} & \multicolumn{3}{c}{Novel-View (2D)} & \multicolumn{4}{c}{Geometry (3D)} \\
            \cmidrule(lr){2-4} \cmidrule(lr){5-8}
            & PSNR$\uparrow$ & SSIM$\uparrow$ & LPIPS$\downarrow$ & CD/diag$\downarrow$ & F@1\%$\uparrow$ & F@2\%$\uparrow$ & F@5\%$\uparrow$ \\
            \midrule
            GART* & 15.898 & 0.848 & 0.199 & 0.0256 & 0.237 & 0.519 & 0.879 \\
            D-3DGS* & \second{20.050} & \second{0.903} & \second{0.103} & \second{0.0181} & \second{0.358} & \second{0.683} & \second{0.952} \\
            \rowcolor{gray!15} Ours & \best{24.027} & \best{0.938} & \best{0.065} & \best{0.0156} & \best{0.439} & \best{0.760} & \best{0.959} \\
            \bottomrule
            \end{tabular}
        }
    \end{subtable}
    \hfill 
    \begin{subtable}[t]{0.42\linewidth}
        \centering
        \caption{Ablation study on Artemis.}
        \label{tab:ablation}
        \vspace{-7pt} %
        \resizebox{\linewidth}{!}{
            \begin{tabular}{lccc}
            \toprule
            \textbf{Method} & PSNR$\uparrow$ & SSIM$\uparrow$ & LPIPS$\downarrow$ \\
            \midrule
            w/o Deform & 22.796 & 0.929 & 0.072 \\
            w/o Shape Prior & 23.891 & 0.937 & 0.068 \\
            w/o Part Anchors & 23.348 & 0.927 & 0.086 \\
            w/o Symm. Enc. & 23.873 & 0.937 & 0.068 \\
            Fixed Skinning & \second{23.934} & \best{0.938} & \second{0.067} \\
            \rowcolor{gray!15} Full & \best{24.027} & \best{0.938} & \best{0.065} \\
            \bottomrule
            \end{tabular}
        }
    \end{subtable}

\label{fig:all_tables}
\end{table}

\subsubsection{Evaluation Protocol.}
We adopt standard ground-truth metrics for evaluating input-view reconstruction. For generation-based methods (GVFDiffusion, DreamMesh4D) that do not produce camera-aligned outputs, we rotate the generated model and select the viewpoint with the highest PSNR for input-view evaluation, providing an upper bound on their input-view performance.
For novel-view quality, we follow SV4D\cite{xie2025sv4d} and use FVD-F (temporal coherence at a fixed view) and FVD-Diag (spatio-temporal consistency). We introduce KID-V (\cref{fig:kid}), a novel-view variant of KID~\cite{binkowski2018kid}, to evaluate novel-view quality (details in supp.), replacing FVD-V for two key reasons. First, FVD-V requires equal frame and view counts, biasing evaluation toward longer videos. Second, FVD-V primarily measures temporal fluency; however, videos rendered by orbiting explicit 3D representations at a fixed timestep are inherently smooth by construction, making FVD-V redundant with per-frame metrics. Instead, KID-V uniformly samples novel views to provide an informative, unbiased measure that remains stable even with limited samples per timestep. We compute KID-V on novel test-set views, while FVD-F and FVD-Diag evaluate the entire generated sequence. For Fauna, a single-image method, we report only KID-16V.

\subsubsection{Quantitative Comparison}
\cref{tab:input-views} and \cref{tab:novel-views} report input-view 
and novel-view metrics. Our method consistently outperforms 
all baselines on DAVIS and Online across both settings.
DreamMesh4D performs best on APTv2 input views, where its strong generative prior compensates for the sparse 15-frame sequences with limited camera variation, while our method ranks second across all metrics.

\begin{figure}[tb]
  \centering
    \includegraphics[width=1.0\linewidth]{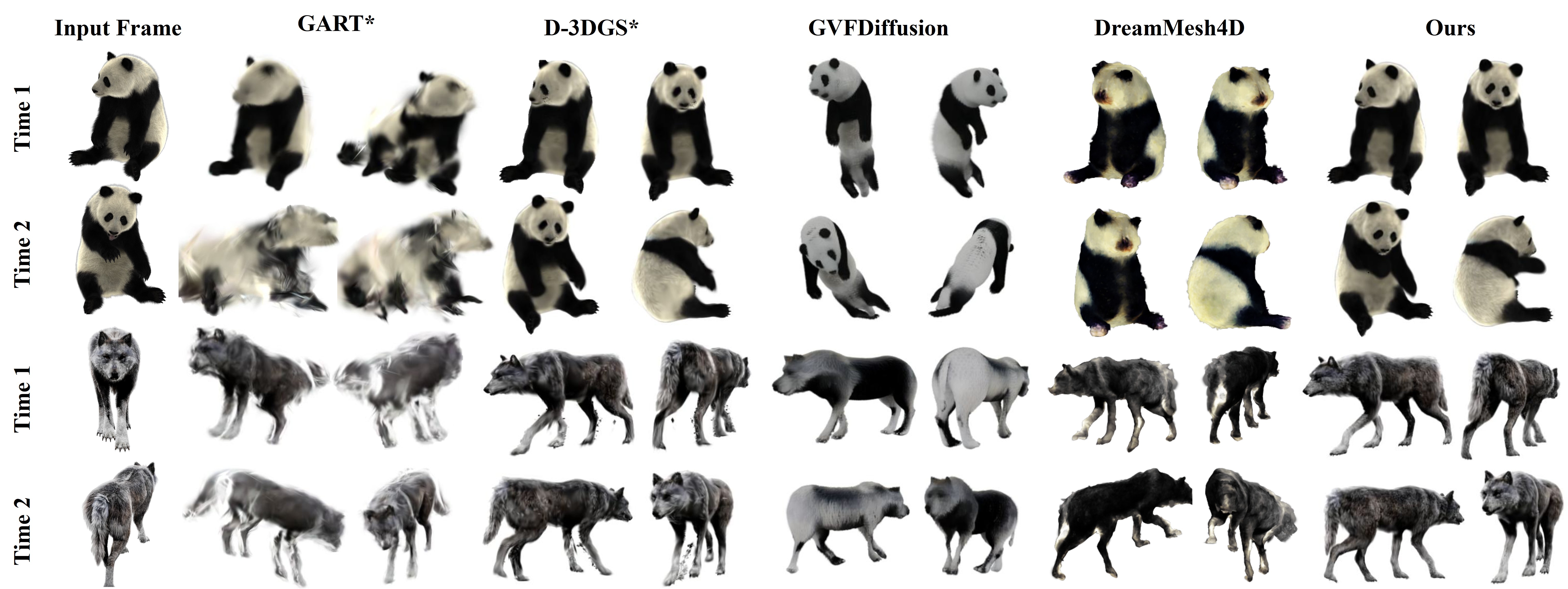}
    \caption{
        Visual comparison with SOTA methods on Artemis at 2 different time steps.
    }
    \label{fig:compare_artemis}
\end{figure}

\cref{tab:artemis_compare} reports quantitative results on the synthetic dataset Artemis, which provides multi-view RGBA renderings and rigged canonical voxel geometry.
The 2D metrics evaluate novel-view rendering quality. 
For 3D geometric evaluation, we derive ground-truth surface points from the canonical voxels via 6-neighbor surface filtering and warp them per frame using the provided skeletal rig (10k points per frame). Predicted Gaussian centers are aligned to the ground truth via 7-DOF similarity ICP; 
we then report Chamfer distance and F-score, both normalized by the per-sequence rest-pose bounding-box diagonal.
Our method outperforms both D-3DGS* and GART* by a substantial margin across all metrics. GVFDiffusion and DreamMesh4D are excluded from this table: their pipelines do not expose camera parameters, precluding the pixel-aligned 2D evaluation; moreover, as their outputs deviate substantially from the input identity, 3D geometric comparison would conflate generation fidelity with reconstruction accuracy. We instead provide a qualitative comparison with all methods in \cref{fig:compare_artemis}.

\begin{figure}[tb]
  \centering
    \includegraphics[width=1.0\linewidth]{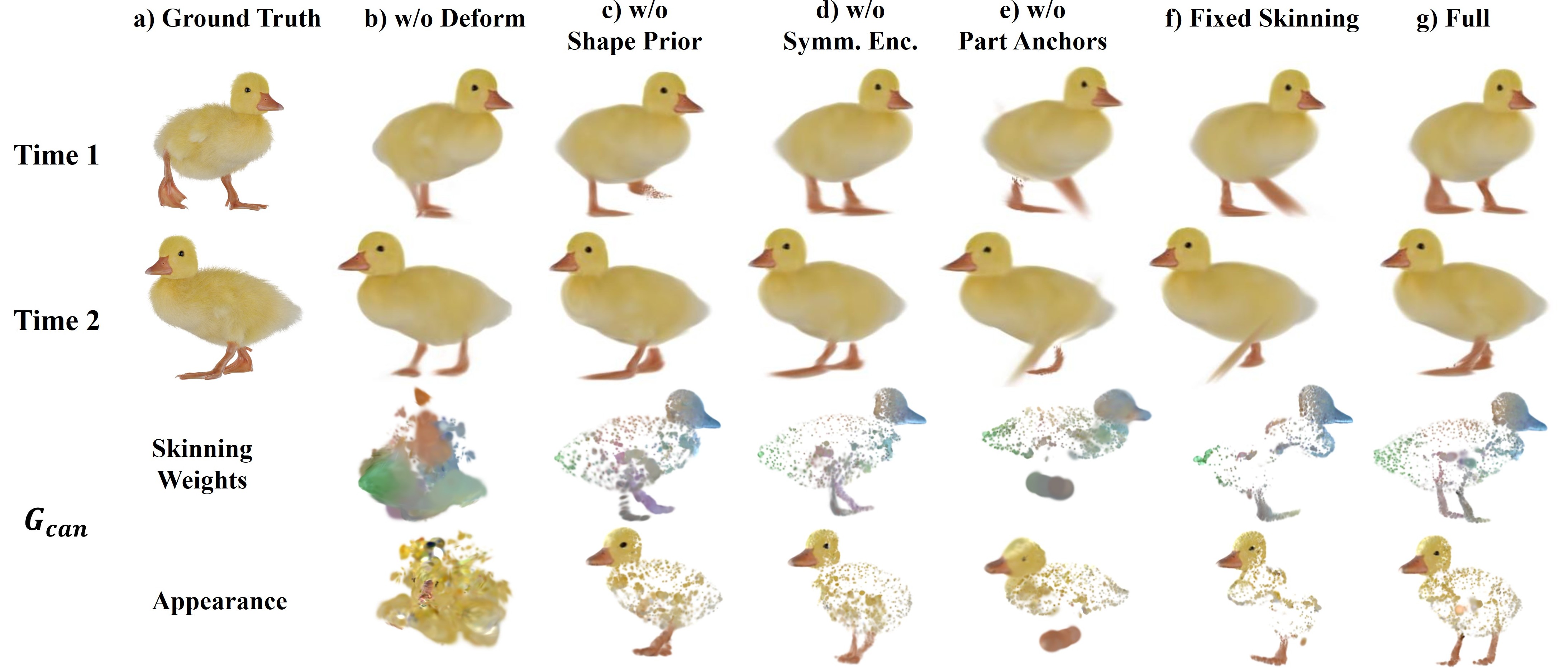}
    \caption{
    Ablation study on Artemis. Top two rows: rendered frames at two time steps. Bottom two rows: canonical Gaussians $\mathcal{G}_{can}$
     colored by skinning weights and appearance. 
    }
    \label{fig:ablation}
\end{figure}

\begin{figure}[tb]
  \centering
    \includegraphics[width=1.0\linewidth]{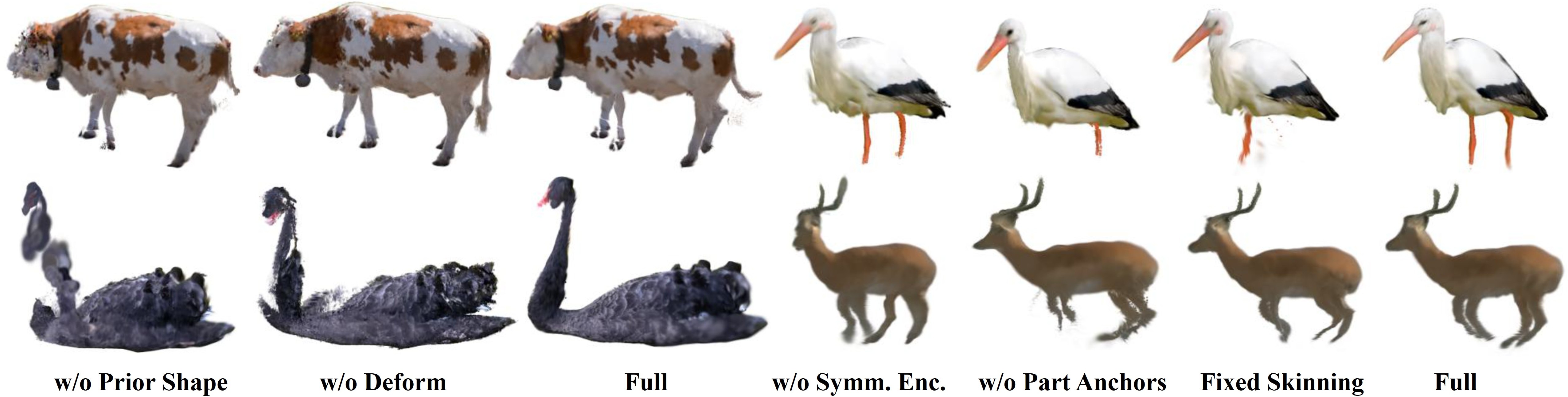}
    \caption{
    Novel-view synthesis under ablations, showing degraded reconstructions without certain components, while the full model remains stable and accurate.
    }
    \label{fig:ablation2}
\end{figure}

\subsubsection{Qualitative Comparison}
As shown in \cref{fig:compare}, baseline methods exhibit various failure modes: incomplete limb geometry (GART*, D-3DGS*), hallucinated structures (GVFDiffusion), and appearance 
inconsistencies (DreamMesh4D). Our method maintains faithful reconstruction across all species.
\cref{fig:compare_artemis} shows comparisons on Artemis. GART* fails on poses that deviate from its D-SMAL template (e.g., the sitting panda) but performs reasonably when the body plan matches (wolf). D-3DGS* exhibits floating artifacts and misses fine motion. GVFDiffusion and DreamMesh4D both fail to preserve instance-specific appearance, producing outputs that deviate substantially from the input. 
Our method faithfully reconstructs both shape and motion across time.

\subsubsection{Ablation Study}

We validate each component through quantitative metrics (\cref{tab:ablation}) and visualization (\cref{fig:ablation}). Specifically: \textbf{w/o Deform} retains only the pose refinement stage; \textbf{w/o Shape Prior} initializes Gaussian positions randomly instead of from prior mesh vertices; \textbf{w/o Symm.\ Enc.} replaces our symmetry-aware encoding with $\mathrm{emb}(t \cdot v) \oplus v$
; \textbf{w/o Part Anchors} removes the learnable anchors and instead concatenates raw skinning weights onto posed attributes as a part identity substitute; \textbf{Fixed Skinning} freezes the skinning field at its prior-initialized values.
As indicated in \cref{tab:ablation}, our full model achieves the best performance across all metrics, with the deformation module and part anchors driving the most significant gains in perceptual quality. 

The canonical-space visualizations (\cref{fig:ablation}) expose the structural root causes of these performance drops.
Without the deformation module (\cref{fig:ablation}b), the model is forced to absorb non-rigid dynamic variations directly into the canonical space, producing a disordered $\mathcal{G}_{can}$. While the final rendering may appear plausible at a coarse glance, subtle limb articulations (note the feet orientation at Time 2) are lost. 
Both removing the shape prior (\cref{fig:ablation}c) and disabling symmetry encoding (\cref{fig:ablation}d) result in an asymmetric canonical shape where one leg is placed in front of the other. Without prior initialization, Gaussians tend to aggregate toward the most frequently observed positions rather than a neutral rest pose. Without symmetry encoding, the model lacks the bilateral cues to recover a symmetric $\mathcal{G}_{can}$. In both cases, reconstruction fails when the target pose reverses the leg ordering relative to the overfitted canonical configuration.
Removing part anchors (\cref{fig:ablation}e) and fixing the skinning field (\cref{fig:ablation}f) produce similar artifacts in the final rendering—only one leg is reconstructed, with the other approximated by stretched Gaussians—but degrade differently in canonical space. In (\cref{fig:ablation}e), the canonical shape is tilted and high-frequency moving parts are collapsed into simplified blobs; in (\cref{fig:ablation}f), the canonical shape itself contains only one leg. Both variants force the deformation stage to hallucinate the missing limb, confirming that learnable part-aware modeling is essential for faithful reconstruction. 
In contrast, the full model (\cref{fig:ablation}g) recovers a well-structured canonical representation and uniquely captures both correct articulated pose and non-rigid deformation of the duck's feet.

We further visualize ablation results on in-the-wild videos in \cref{fig:ablation2}. Removing the shape prior leads to severe structural collapse, while removing the deformation module preserves overall structure but introduces artifacts and fails to capture fine motion such as the cow's tail swing.
Disabling symmetry encoding leads to blurred or distorted geometry in real videos with camera estimation noise. Without part anchors or with a fixed skinning field, limb reconstruction degrades noticeably, confirming their importance for unconstrained settings.

\begin{figure}[tbp]
\centering
    \begin{subfigure}[b]{0.22\linewidth}
        \centering
        \includegraphics[width=\linewidth]{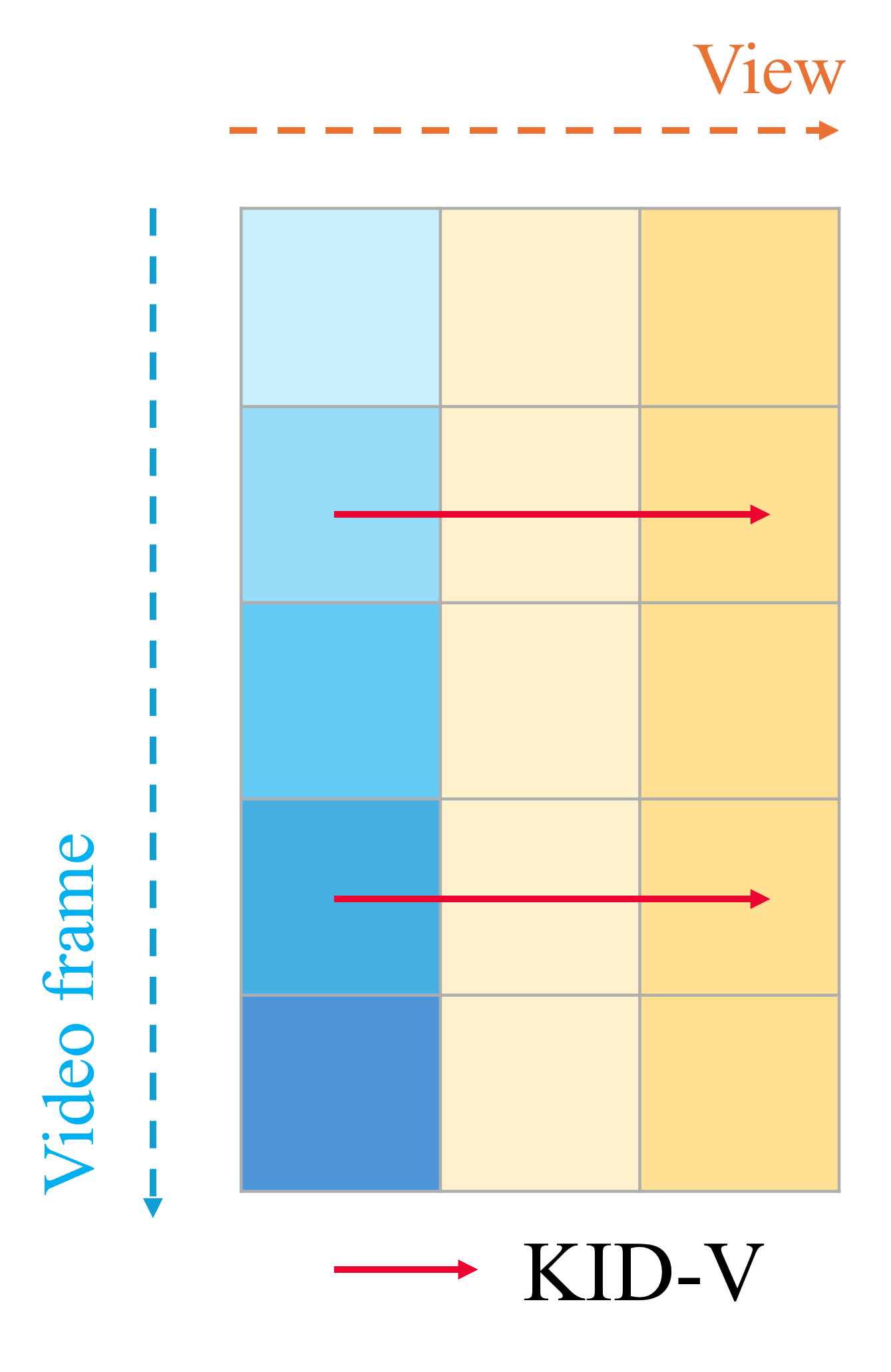}
        \caption{KID-V metric schematic.}
        \label{fig:kid}
    \end{subfigure}
    \hfill 
    \begin{subfigure}[b]{0.75\linewidth}
        \centering
        \includegraphics[width=\linewidth]{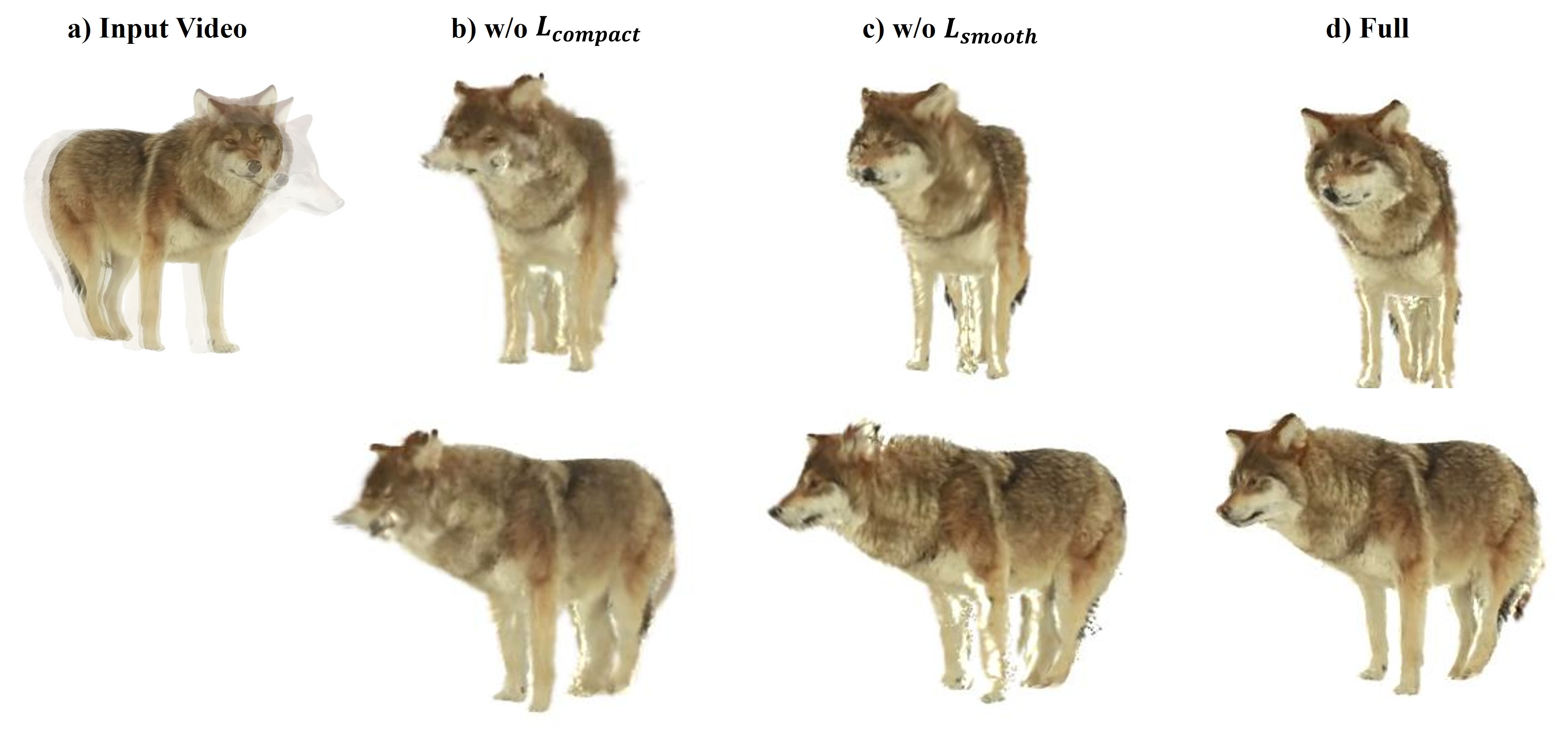}
        \caption{Geometric regularization ablation.}
        \label{fig:ablation_reg}
    \end{subfigure}

\caption{(a) Conceptual illustration of our proposed KID-V metric. 
(b) Visual ablation of $\mathcal{L}_{\text{compact}}$ and $\mathcal{L}_{\text{smooth}}$. 
    Without $\mathcal{L}_{\text{compact}}$, the head splits into duplicated fragments.
    Without $\mathcal{L}_{\text{smooth}}$, the legs and head exhibit rough, noisy surfaces. 
    The full model maintains coherent geometry with clean surfaces across viewpoints.}
\label{fig:kid_geo}
\end{figure}

\subsection{Discussion and Limitations}

While our method performs well across diverse species, several limitations remain.

\noindent\textbf{Robustness to Unstable Camera Estimation.}
Our filtering strategy processes the original and flipped sequences independently. A frame discarded in one sequence due to unstable camera estimation can still be optimized through its counterpart in the other sequence via our bilateral augmentation. Nevertheless, under severe camera instability where both sequences are affected, optimization may converge to a compromised solution.

\noindent\textbf{Imperfect Mask Supervision.}
We directly use Grounded-SAM masks without manual refinement; inaccurate segmentation reduces visual supervision and degrades geometry in affected regions. Side-to-side self-occlusion is partially mitigated by our bilateral augmentation, which supplies supervision from the mirrored view, while persistent occlusion across the sequence remains a limitation.

\noindent\textbf{Ambiguous Localized Motion.}
Head motion remains difficult under limited viewpoints. Unlike limbs, head rotations produce only subtle silhouette changes, making silhouette supervision insufficient. Although our geometric regularization alleviates head splitting (\cref{fig:ablation_reg}), it cannot fully resolve this ambiguity.

\noindent\textbf{Evaluation Limitations.}
Rendering-based metrics are indirect proxies for geometric accuracy: high rendering scores 
do not guarantee correct articulation. On Artemis, the geometric metrics provide a more direct assessment, but for in-the-wild videos ground-truth 3D geometry is unavailable, leaving rendering-based proxies as the only option.

Future work includes improving robustness to unreliable camera and mask estimates, incorporating stronger geometric or structural priors, and developing more reliable evaluation protocols.

\section{Conclusion}
We presented a progressive test-time optimization framework 
for 4D animal reconstruction from monocular video. By 
treating shape priors as coarse initialization with a 
learnable skinning field, introducing symmetry-aware temporal 
encoding to exploit bilateral cues under camera noise, and 
conditioning non-rigid deformation on part anchors shared 
with the pose estimation stage, our method disentangles 
articulated motion from non-rigid deformation without 
requiring multi-view supervision or category-specific 
templates. Extensive experiments across diverse species 
demonstrate improvements over existing baselines in 
reconstruction quality and temporal consistency. While 
developed for animals, the principle of balancing prior 
knowledge with optimization flexibility may extend to other 
articulated non-rigid objects.


%
%
\bibliographystyle{splncs04}
\bibliography{main}



\end{document}